\documentclass[letterpaper, 10 pt, conference]{ieeeconf}  

\IEEEoverridecommandlockouts                           
\usepackage{microtype}
\usepackage{amsmath,amssymb,amsfonts}
\usepackage{algorithmic}
\usepackage{graphicx}
\usepackage{textcomp}
\usepackage{xcolor}
\usepackage{subfig}
\usepackage{gensymb}
\usepackage{caption}
\usepackage{hyperref}
\usepackage{mwe}
\usepackage{amsfonts}
\usepackage{amssymb}
\usepackage{verbatim}
\usepackage{amsbsy}
\usepackage{siunitx}
\usepackage{tabularx, booktabs}
\usepackage{dblfloatfix}
\usepackage{cite}
\usepackage{cleveref}

\usepackage{algorithm, algorithmic}
\usepackage[autolanguage]{numprint}

\usepackage{spreadtab}
\usepackage{multirow}
\usepackage{cite}
\usepackage{hyperref}
\usepackage{xcolor}
\usepackage{algorithmic}
\usepackage{textcomp}
\usepackage{xcolor}
\usepackage{subfig}
\usepackage{gensymb}
\usepackage{caption}
\usepackage{mwe}
\usepackage{amsfonts}
\usepackage{amssymb}
\usepackage{verbatim}
\usepackage{amsbsy}
\usepackage{siunitx}
\usepackage{tabularx, booktabs}
\usepackage{soul}
\usepackage{tikz}
\usetikzlibrary{calc}

\makeatletter
\newif\if@anonymize

\@anonymizetrue    

\if@anonymize
  \newcommand{\highlight@DoHighlight}{
    \fill [outer sep = -15pt, inner sep = 0pt, color=black]
          ($(begin highlight)+(0,8pt)$) rectangle ($(end highlight)+(0,-3pt)$) ;
  }

  \newcommand{\highlight@BeginHighlight}{
    \coordinate (begin highlight) at (0,0) ;
  }

  \newcommand{\highlight@EndHighlight}{
    \coordinate (end highlight) at (0,0) ;
  }

  \newdimen\highlight@previous
  \newdimen\highlight@current
  \newlength{\item@width}

  \DeclareRobustCommand*\anonymize{%
    \SOUL@setup
    \def\SOUL@preamble{%
      \begin{tikzpicture}[overlay, remember picture]
        \highlight@BeginHighlight
        \highlight@EndHighlight
      \end{tikzpicture}%
    }%
    \def\SOUL@postamble{%
      \begin{tikzpicture}[overlay, remember picture]
        \highlight@EndHighlight
        \highlight@DoHighlight
      \end{tikzpicture}%
    }%
    \def\SOUL@everyhyphen{%
      \discretionary{%
        \SOUL@setkern\SOUL@hyphkern
        \SOUL@sethyphenchar
        \tikz[overlay, remember picture] \highlight@EndHighlight ;%
      }{%
      }{%
        \SOUL@setkern\SOUL@charkern
      }%
    }%
    \def\SOUL@everyexhyphen##1{%
      \SOUL@setkern\SOUL@hyphkern
      \settowidth{\item@width}{##1}%
      \makebox[\item@width]{}%
      \discretionary{%
        \tikz[overlay, remember picture] \highlight@EndHighlight ;%
      }{%
      }{%
        \SOUL@setkern\SOUL@charkern
      }%
    }%
    \def\SOUL@everysyllable{%
      \begin{tikzpicture}[overlay, remember picture]
        \path let \p0 = (begin highlight), \p1 = (0,0) in \pgfextra
          \global\highlight@previous=\y0
          \global\highlight@current =\y1
        \endpgfextra (0,0) ;
        \ifdim\highlight@current < \highlight@previous
          \highlight@DoHighlight
          \highlight@BeginHighlight
        \fi
      \end{tikzpicture}%
      \settowidth{\item@width}{\the\SOUL@syllable}%
      \makebox[\item@width]{}%
      \tikz[overlay, remember picture] \highlight@EndHighlight ;%
    }%
    \SOUL@
  }
\else
  \newcommand{\anonymize}[1]{#1}
\fi
\makeatother 

\title{\LARGE \bf
Active Perception Applied To Unmanned Aerial Vehicles Through Deep Reinforcement Learning
}

\author{Matheus G. Mateus$^{1}$, Ricardo B. Grando, Paulo L. J. Drews-Jr$^{2}$
\thanks{$^{1}$Matheus G. Mateus, P. L. J. Drews-Jr are with Centro de Ci\^encias Computacionais (C3) of  Universidade Federal do Rio Grande (FURG), RS, Brazil. E-mail: {\tt\small paulodrews@furg.br}}
\thanks{$^{2}$Ricardo B. Grando is with the Universidad Tecnol\'ogica del Uruguay (UTEC). E-mail: {\tt\small ricardo.bedin@utec.edu.uy}}
}

\begin{document}

\maketitle
\thispagestyle{empty}
\pagestyle{empty}

\begin{abstract}
Unmanned Aerial Vehicles (UAV) have been standing out due to the wide range of applications in which they can be used autonomously. However, they need intelligent systems capable of providing a greater understanding of what they perceive to perform several tasks. They become more challenging in complex environments since there is a need to perceive the environment and act under environmental uncertainties to make a decision. In this context, a system that uses active perception can improve performance by seeking the best next view through the recognition of targets while displacement occurs. This work aims to contribute to the active perception of UAVs by tackling the problem of tracking and recognizing water surface structures to perform a dynamic landing. We show that our system with classical image processing techniques and a simple Deep Reinforcement Learning (Deep-RL) agent is capable of perceiving the environment and dealing with uncertainties without making the use of complex Convolutional Neural Networks (CNN) or Contrastive Learning (CL).
\end{abstract}



\section{Introduction}
\label{introduction}


Unmanned aerial vehicles (\textit{UAVs}) are one of the types of autonomous vehicles that have been gaining more and more space due to the versatility in their possible applications. Some of them face problems that involve acting on the water to perform their respective tasks, in the different areas of hydrological monitoring~\cite{velez2021applications}, mainly in remote sensing applications~\cite{perumal2017uav}. A problem that also arises in search and rescue missions with the recognition of victims on the water surface~\cite{qingqing2020towards}, joint navigation between autonomous boats and \textit{UAVs}~\cite{garberoglio2019coordinated} and reconnaissance and pollution monitoring by detecting litter on the sea surface~\cite{gonccalves2020mapping}. The challenges of detecting, tracking, and following a static or moving target of interest are critical for these applications. The recognition of these targets becomes much more complex compared to acting on solid terrains, considering that the reflectance of light suffered by ripples on the water surface and other different influences of the incidence of light impairs the optical flow captured in the images during the flight.

The need to combine the information contained in the images and in the reading of other sensors during the flight for the decision-making of the vehicle is often a complex problem since the readings are directly affected by the current positioning and the future positioning affected by the current reading. Thus, bringing the demand to perform the movement, always seeking the greatest gain of information or the next best view (\textit{NBV}), and thus properly characterizing an active perception~\cite{gemerek2020active}.


In this context, this work proposes to deal with the problem of acting on the water by performing tracking and displacement, actively perceiving the environment for the completion of the landing on a base arranged on the surface of the water. We adopted the \textit{AirSim} plugin of \textit{Unreal Engine}~\cite{airsim2017fsr}, aiming to use the photorealistic environments, combined with a Deep Reinforcement Learning approach (\textit{Deep-RL}) through the support provided to this type of approach. With the analysis of the \textit{Deep-RL} approach, we found that the algorithms tend not to converge with long observation input, where the use of images as a means of observing the agent corresponds to this problem. By simplifying the processing, we focus on converting the raw image data in the distances in \textit{pixels} referring to the axes $x$ and $y$ to the target. Thus, leading the algorithm to converge without the requirement of complex convolutional neural networks (CNN) or contrastive learning (CL) even within a highly realistic environment, allowing its usage in real-world embedded systems.

This work contains the following main contributions:

\begin{itemize}

\item Realistic simulation using the plugin \textit{AirSim} with the application of the algorithm of \textit{Deep-RL} \textit{Deep Deterministic Policy Gradients - DDPG} using an \textit{encoder} in order to solve the high dimension space problem of observations. 

\item Active vision system based on \textit{Deep-RL} and image processing for allowing autonomous landing in a boat, a challenging environment.
\end{itemize}

The organization of this work is given as follows: Section II discusses the related works, followed by the presentation of the proposed active vision method in Section III. Next, the  results are presented in Section IV. Finally, the conclusions are future work are drawn in Section V.

\section{Related Work}
\label{related_works}

The concept of an active observer, pre-established both by~\cite{bajcsy2018revisiting} and by~\cite{gemerek2020active}, is given when the observer performs some type of activity whose purpose is to control the geometric parameters of the sensory apparatus. The purpose is to manipulate the constraints underlying the observed phenomena to improve the quality of the perceptual results. That is, active perception occurs when an observer changes its position and/or the position of its sensors in order to obtain the greatest amount of information about the environment in which it is inserted. Also defined as the union of intelligent control or active control and visual perception or active vision.

This concept is much explored in applications that seek to somehow reconstruct the environment or the scene itself, as in ~\cite{kong2020active}and\cite{nishimura2018active}, where the estimation of \textit{NBV} is a little more critical in optimization terms. In situations closer to real-world applications, some adaptations occur, such as in relation to the positioning of the visual sensor, but the principle is still the same as in~\cite{liu2012fast},~\cite{danelljan2014low} and~\cite{czuni2018lightweight}. There are mainly approaches related to the problem of flying over and detecting targets on the surface of the water with the displacement, such as the recognition and differentiation of terrain~\cite{pombeiro2015water} and ~\cite{ matos2018uav} and the search for specific targets on the surface like em~\cite{qingqing2020towards} and ~\cite{singh2020simulating}.

With the analysis of these works, it is possible to perceive the use of several algorithms to estimate and carry out the positioning of the UAV. In order to take advantage of the support of \textit{AirSim} and given the broad concept of active perception in relation to the algorithms and their estimates, references were searched for using \textit{Deep-RL} algorithms focused on displacement based on the visual perception. Many works were analyzed where it was proposed to use the algorithm of \textit{Deep-RL} called Deep Deterministic Policy Gradients (\textit{DDPG}) applied to the search and rescue task in a closed environment. This same algorithm was used by \cite{kang2019generalization}, focusing on recognizing and avoiding targets. It was also approached by \cite{rodriguez2019deep} where the use of \textit{DDPG} was proposed to perform the landing on a mobile base. The three cited references make use of both simulated data (Gazebo) and real data, in addition to validating the method in both environments. Differently from them, we focus on an active perception approach based on Deep-RL to achieve autonomous landing based on the vision in a boat, a challenging target.

\section{Methodology}
\label{methodology}

\subsection{Simulation Environment}
The choice of plugin \textit{AirSim} was made seeking a significant improvement in the quality of the environment in which the UAV would act. The simulator is developed aiming at the costs of developing and testing algorithms for autonomous vehicles in the real world. \textit{AirSim} provides an interface to define the vehicle as a rigid body that can have an arbitrary number of actuators generating forces and torques. The generation of all mathematical models focused on vehicle physics and its sensors was estimated based on real vehicles, bringing a very close result between the simulation and the performance of a real UAV as described by~\cite{shah2018airsim}. \textit{AirSim} was developed for the \textit{Unreal Engine} (https://www.unrealengine.com), being developed and maintained by the game developer \textit{Epic Games} which is responsible for the development of several famous games in the market. The great quality presented in the graphics provided by the engine has been bringing new horizons to its use, were on its own platform. In this way, the use of simulation for real vehicles has been growing and having implications as flight and driving simulators where the similarity with a real environment is of high importance.

\begin{figure}[thpb]
      \centering
      \includegraphics[scale=0.2]{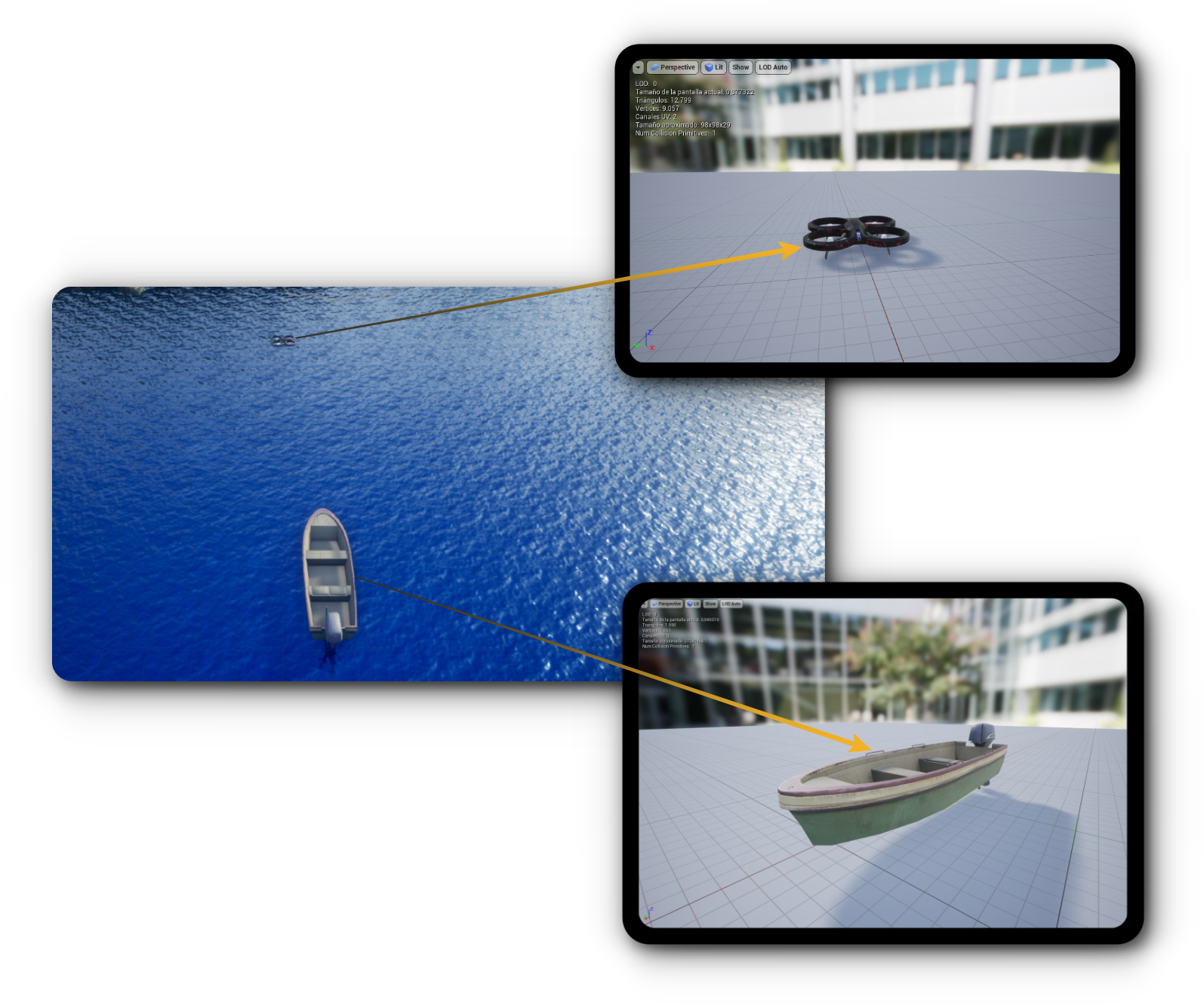}
      \caption{Sample images of the simulation environment composed of a lake, boat(target), and a drone.}
      \label{fig_ambiente_unreal}
\end{figure}

In this work, the engine version 4.25 was adopted together with the environment called \textit{Landscape Mountain} provided by the  \textit{engine} development market, along with a water surface (https://free3d.com/3d- model/ocan-1761.html) and a fishing boat (https://free3d.com/3d-model/boat-6219.html) shown in Fig~\ref{fig_ambiente_unreal}. The vehicle used by \textit{AirSim} is a model \textit{AirDrone} with a wide range of possible sensors that can be added to its structure, such as \textit{GPS}, Lidar, magnetometer, barometer and more than one camera with some image formats beyond the conventional \textit{RGB}. These settings can all be modified through a \textit{.xml} file with the respective parameters of each sensor. Thus, in the present work, only the barometer and a camera were used with the conventional capture allocated in the center of the vehicle directed downwards in the same way as in~\cite{liu2012fast}.

\subsection{Active Perception System}
A ROS package (https://www.ros.org/) called \textit{"active\_perception"} was developed, where two nodes are implemented, respectively seeking detection and landing. The node in charge of detection is called \textit{"cam\_listener"}, responsible for reading the camera and publishing in the topic \textit{"vision"} the message with the target distances in pixels on the axes $x$ and $y$. And the \textit{landing} node is responsible for the displacement decisions and the proper landing based on the readings of the \textit{"vision"} topic. The code will be freely available in case of acceptance.

\subsubsection{Image Detection}
In order to detect the boat in a simple and efficient way, the search is divided into two steps. Firstly, we analyze the reflection of light on the surface of the water and then the search for the boat. For the light detection, the original image captured by the camera is converted to grayscale and then Gaussian Blur (1) is applied, evaluating the variance of pixels in both directions and saving the first processed frame. From the second frame, the optical flow in the image is analyzed using the Lucas-Kanade~\cite{baker2004lucas} method. A comparison is made in the displacement of pixels based on the distance from the neighbors, looking for changes having the parameters adapted for the reflection of the light.

$$
G(x, y) = \frac{1}{2\pi\sigma^2}\dot{e}^{\frac{x^2 + y^2}{2\sigma^2}} \eqno{(1)}
$$

\begin{figure}[htb]
    \begin{minipage}[c][0.75\width]{0.235\textwidth}
	   \centering
	   \includegraphics[width=\textwidth]{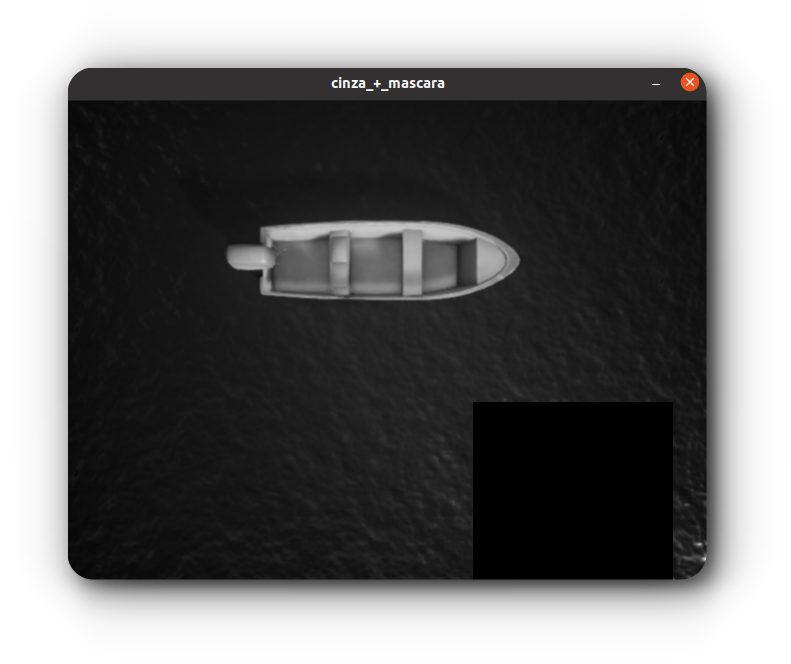}
          \textbf{a)}
	\end{minipage}
    \begin{minipage}[c][0.75\width]{0.235\textwidth}
	   \centering
	   \includegraphics[width=\textwidth]{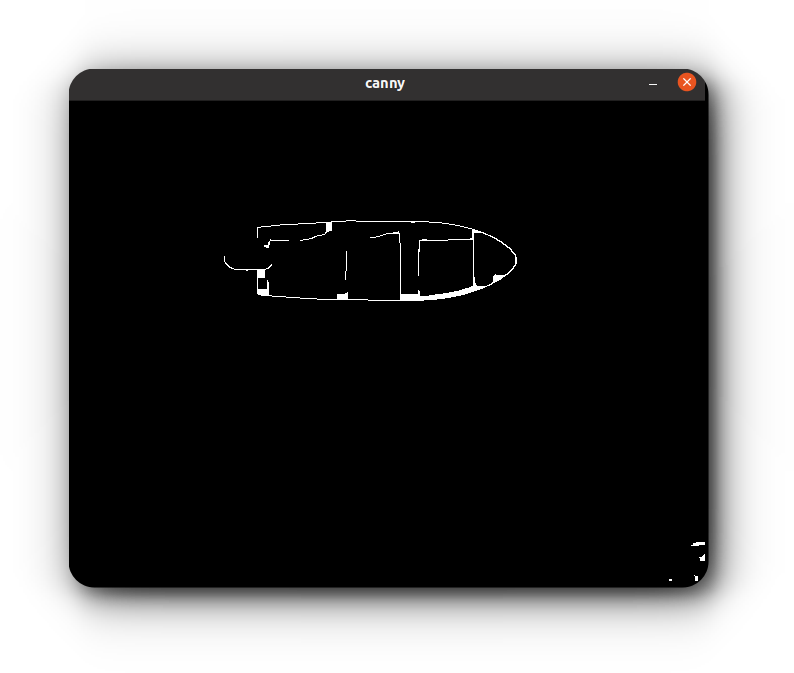}
          \textbf{b)}
	\end{minipage} \vfill
    \begin{minipage}[c][1.15\width]{0.235\textwidth}
	   \centering
	   \includegraphics[width=\textwidth]{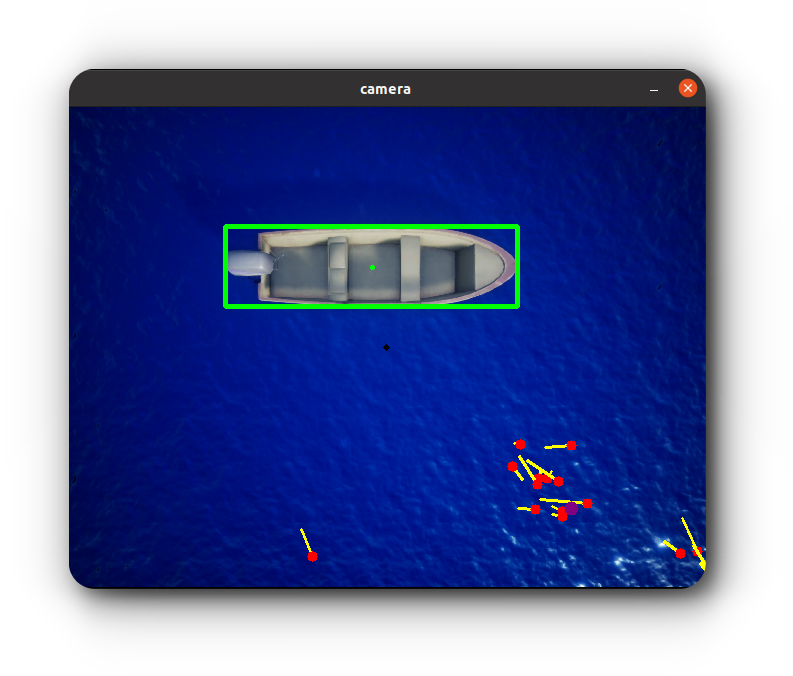}
          \textbf{c)}
	\end{minipage}
 \caption{Agent's obtained scene view through the cam.}
      \label{fig_filters}
\end{figure}

Based on the points where modifications were found, the centroid of the region generated by the points is calculated, and then a mask is created from it to be applied over the grayscale image as shown in Fig~\ref{fig_filters} (a) . With it, the \textit{Canny} algorithm~\cite{canny} is applied to detect possible edges by analyzing both the width and the length of the image.
%

Finally, a morphological transformation is applied to the image. As a dilation followed by erosion in search of the contour with the largest area, resulting in Fig~\ref{fig_filters} (b) and by calculating the center of the contour, the distances in \textit{pixels} from the target in the axes $x$ and $y$ and published in the topic \textit{"$\backslash$vision"}. Thus, visually the result of the two analyzes happens as shown in Fig~\ref{fig_filters} (c) with the highlighting of the target and the visualization of the reflection on the surface by the optical flow analysis.

\subsubsection{Landing}
In the \textit{DDPG}~\cite{silver2014deterministic} algorithm, the actor directly maps states to actions instead of generating the probability distribution in a discrete action space as is originally done by other algorithms of the same model. The other two networks are destination networks, being time-delayed copies of their original networks that smoothly track the learned networks. The use of target value networks greatly improves stability in learning, since the main network update equations are dependent on the values calculated by themselves making them prone to divergence.

\textit{DDPG} also uses a repeating \textit{buffer} to sample the experience and update the neural network parameters. In reinforcement learning for discrete action spaces, exploration is done through the probabilistic selection of a random action. For continuous action spaces, exploration is done by adding noise to the action itself, as it was used in the work, and can also be used in the state. One of the most used methods is the \textit{Ornstein-Uhlenbeck}~\cite{uhlenbeck1930theory} process to add noise to the output. It is correlated with the previous noise in order to prevent the noise from canceling out the overall dynamics created by the Actor.

\begin{figure}[thpb]
      \centering
      \includegraphics[scale=0.165]{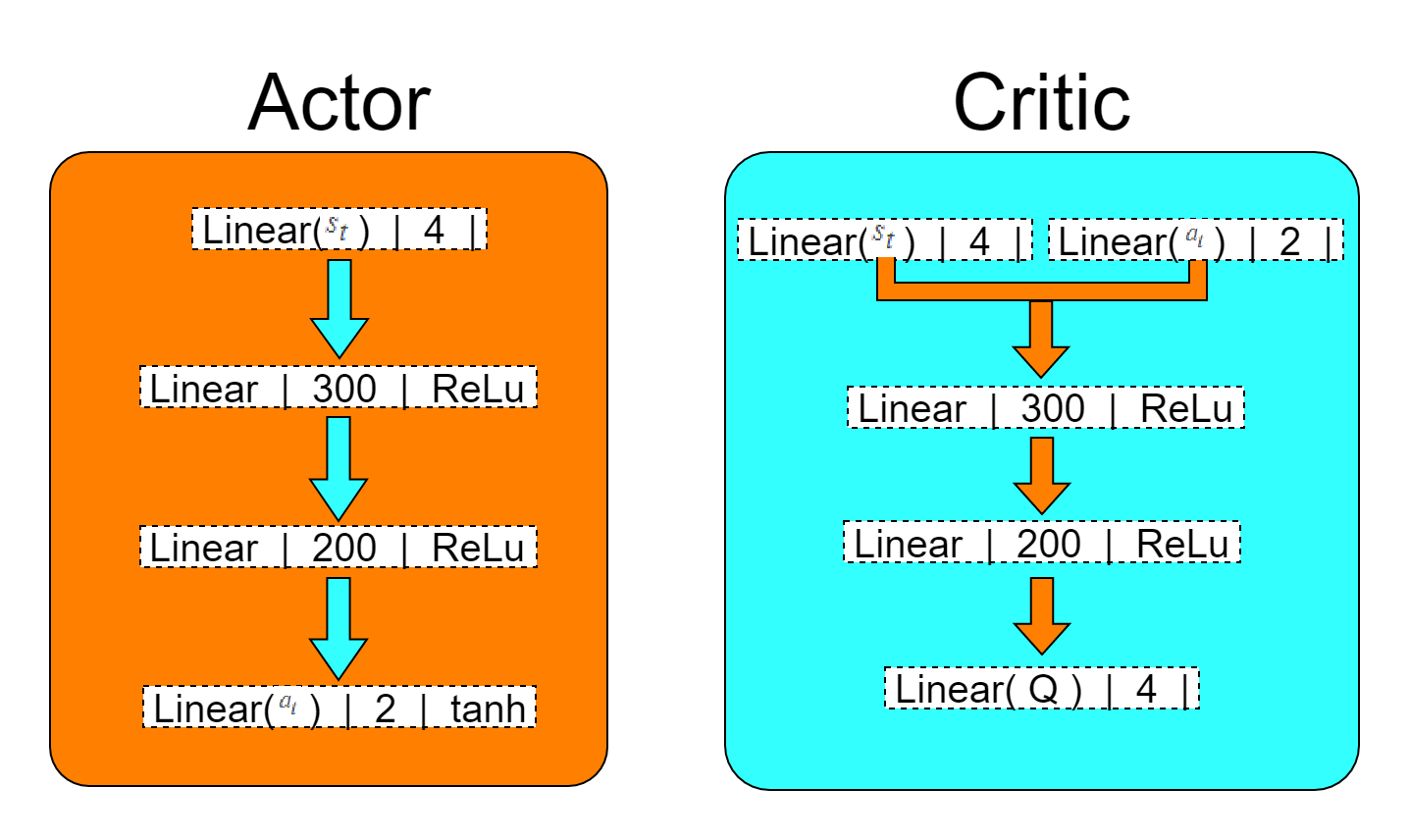}
      \caption{Actor and Critic neural networks.}
      \label{fig_redes}
\end{figure}

Our model is built using a similar structure of the network \cite{rodriguez2019deep} as shown in Fig.~\ref{fig_redes}, but changed the dimensions of the hidden layers to 300 and 200 units due to the simplicity proposed for the inputs.

For the action space, the same principle was maintained, using only the linear velocities in $x$ and $y$ with the velocity in $z$ being constant. In the state space, the $x$ and $y$ distances in pixels read through the publication in the topic \textit{"vision"} made with the image processing are passed to the agent. For the implication of the actions, considering that the output of the Ator network varies between -1 and 1 because it is a hyperbolic tangent function, after the noise \textit{Ornstein-Uhlenbeck} is increased, normalization between -0.25m/s and 0.25 is performed. m/s to be passed as speeds to the vehicle's low-level control. A reward function was developed where $d_t$ is obtained through the module of distances $x$ and $y$ in pixels (4), resulting in function (5) that evaluates the distances to send the appropriate reward to the agent. Rewards are given based on completion of the landing with the highest reward, a minimum reward for decreasing the distance, no reward for an increase in the distance to the target, and finally a negative reward for when the target is lost from sight.
$$
 d_t = \sqrt{{d}_{x}^{2} + {d}_{y}^{2}} \eqno{(4)}
$$

$$
r(s_t, a_t) = 
              \begin{cases}
                250          & \text{if } d_t <= 10\\
                0.1          & \text{if } d_t < d_{t-1}\\
                0            & \text{if } d_t >= d_{t-1}\\
                -10          & \text{if } d_t = 1000000.0\\
                
            \end{cases} \eqno{(5)}
$$

\begin{figure}[thpb]
      \centering
      \includegraphics[width=\columnwidth]{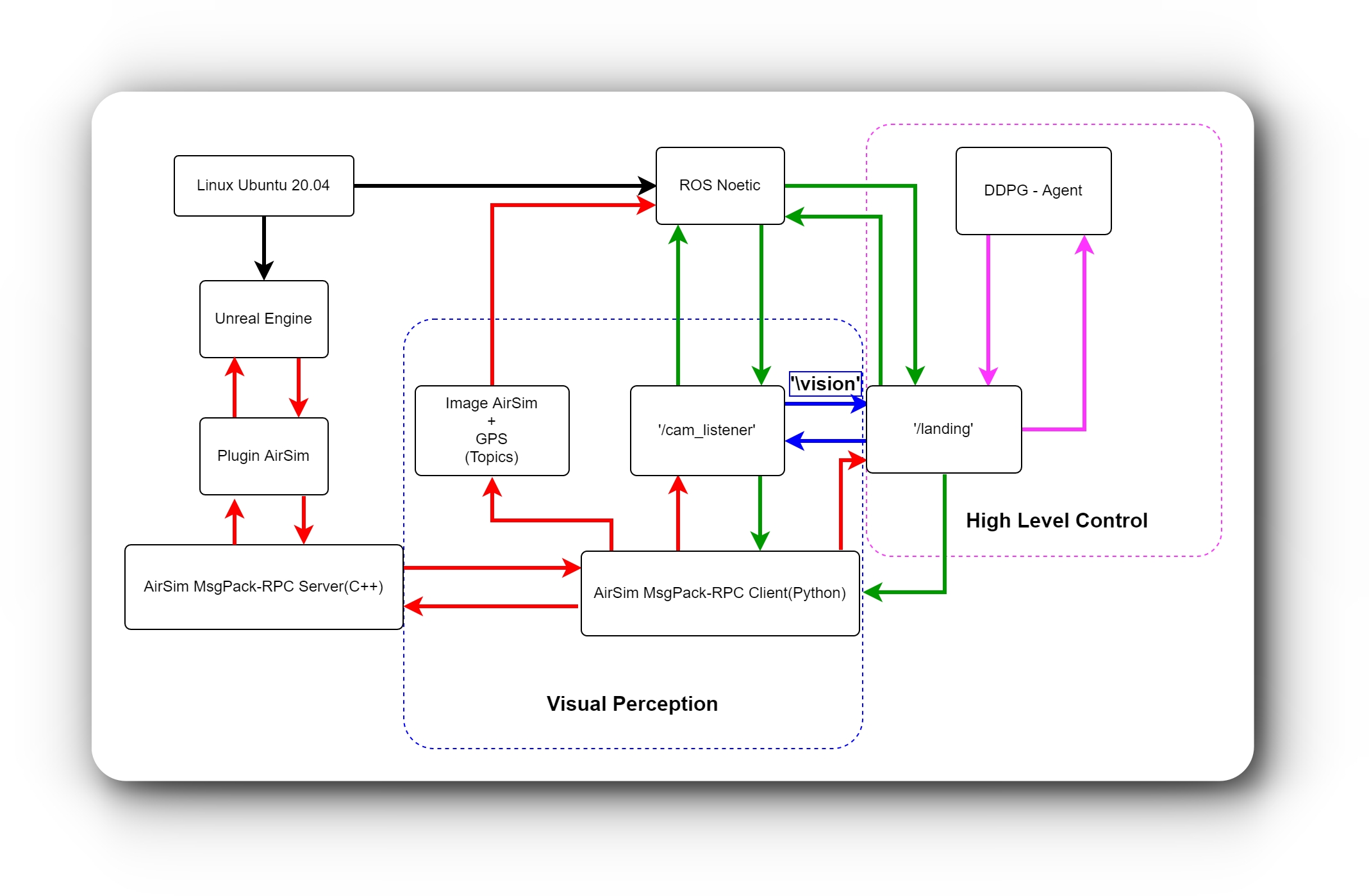}
      \caption{Proposed active perception system.}
      \label{fig_diagrama_sistema}
\end{figure}

The structure of the final system characterizing the active perception is presented in Fig.~\ref{fig_diagrama_sistema}, leading then to the analysis of the results obtained by the algorithm presented in the next section.
\section{Experimental Results}
\label{results} 

The system is implemented using the operating system \textit{Ubuntu 20.04 LTS}, with the language \textit{C++} for image processing in version $9.3.0$ and the language \textit{Python} used for agent training through the \textit{Pytorch} module in version $3.8.10$. Considering the possibility of acceleration by \textit{hardware} (\textit{GPU}) through the \textit{Unreal Engine} and also its demand in graphics processing, a \textit{GPU} model \textit{RTX 2080 Ti} with version $10.1$ supporting CUDA parallel processing is adopted. A \textit{CPU} \textit{Intel Core i7-7700k} is used for the main processing, also counting on 32 \textit{GB} of memory \textit{RAM}.

\subsection{Training}
\begin{figure}[htb]
    \begin{minipage}[c][1.1\width]{0.235\textwidth}
	   \centering
	   \includegraphics[width=1.16\textwidth]{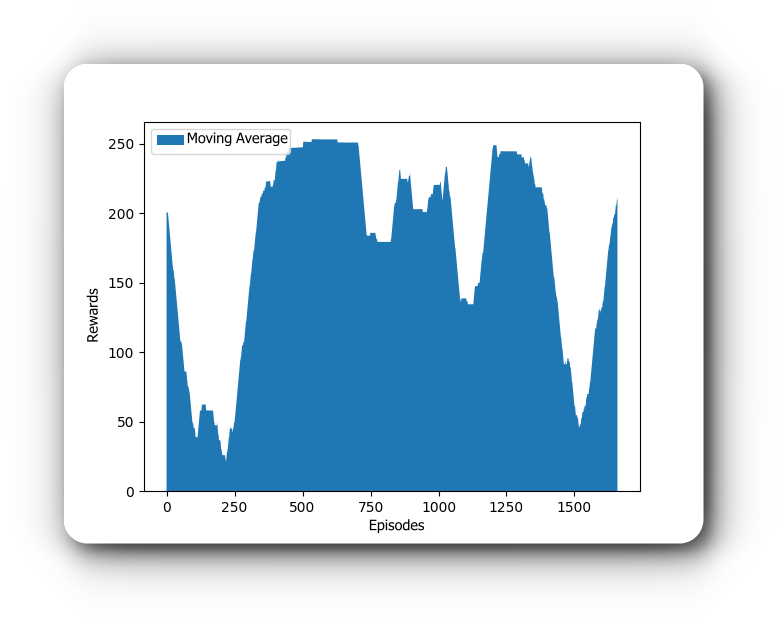}
          \textbf{a)}
	\end{minipage}
    \begin{minipage}[c][1.1\width]{0.235\textwidth}
	   \centering
	   \includegraphics[width=1.16\textwidth]{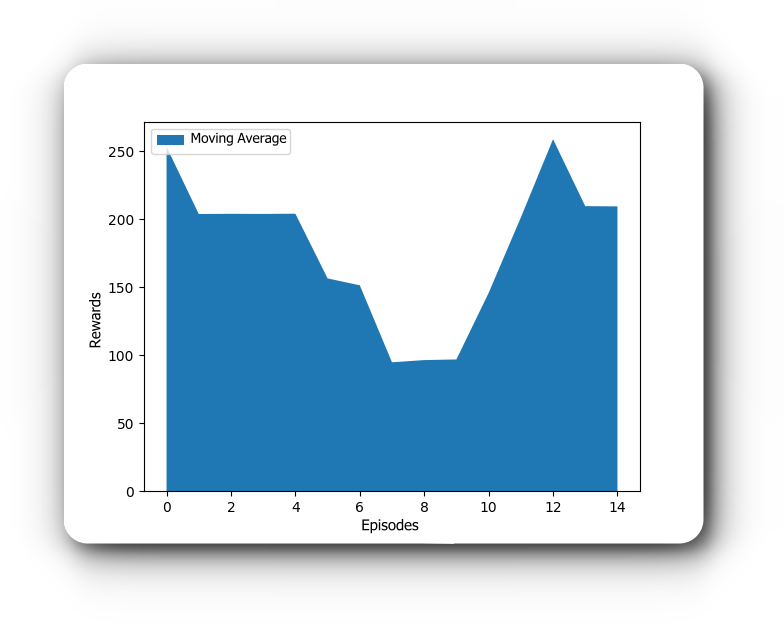}
          \textbf{b)}
	\end{minipage} \vfill
 \caption{a) Contains rewards from episode 0 to episode 1780  and b) Contains rewards from episode 1781 to episode 1800.}
      \label{fig_treino_r}
\end{figure}

\begin{figure}[htb]
    \begin{minipage}[c][1.1\width]{0.235\textwidth}
	   \centering
	   \includegraphics[width=1.16\textwidth]{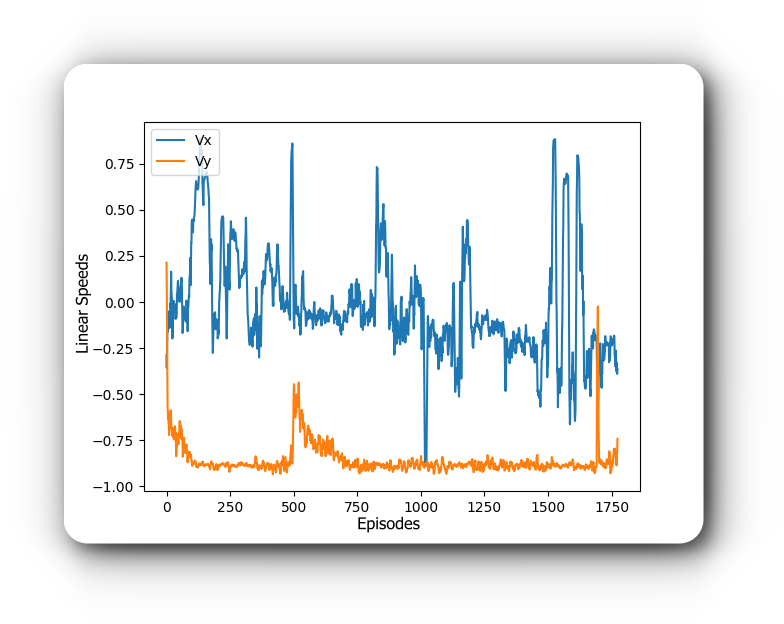}
          \textbf{a)}
	\end{minipage}
    \begin{minipage}[c][1.1\width]{0.235\textwidth}
	   \centering
	   \includegraphics[width=1.16\textwidth]{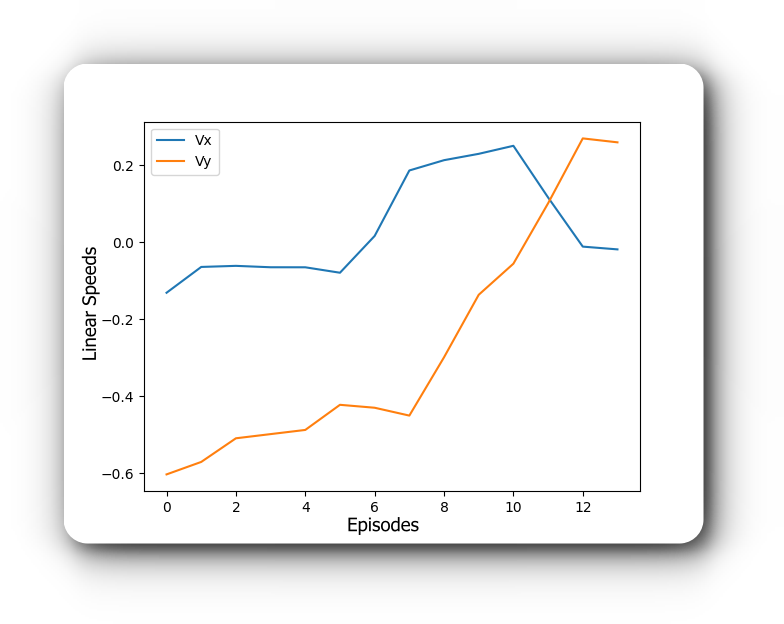}
          \textbf{b)}
	\end{minipage} \vfill
 \caption{a) Contains speeds from episode 0 to episode 1780  and b) Contains speeds from episode 1781 to episode 1800.}
      \label{fig_treino_v}
\end{figure}

For the training phase, the situation imposed on the agent consists of the analysis of the distance to the boat through the reward function previously demonstrated in (5). If the agent gets the maximum reward within an episode, the environment is restarted and the episode ends. For the other rewards, such as if the agent loses sight of the boat and gets a negative reward, only the simulation is restarted. In constant rewards, there is no restart of the simulation, only the attribution of the reward. The agent parameters are defined based on the parameters used by \cite{rodriguez2019deep}, being the state space ($4$), the action space ($2$), the repetition buffer ($50000$), the gamma factor ( $0.99$), the actor-network learning rate ($1e^{-4}$), the critic network learning rate ($1e^{-3}$), and the update factor for the target networks ( $1e^{-3}$). It was defined to use $1000$ steps within each episode, leading to the use of batches of $512$ samples to be saved in memory. $1800$ episodes equivalent to more than 24 hours of training, looking for a policy capable of making the landing. In the first training phase, a target is arranged in a single position, as shown in Fig.~\ref{fig_filters}, and in the second with positions vary randomly in case there is a landing. With the analysis of the results referring to the final part, a significant improvement in the policy generalization learned by the agent was noticed, leading to the conclusion that the learning had been carried out. The figures Fig.~\ref{fig_treino_r} (a) and Fig.~\ref{fig_treino_r} (b) illustrate the moving averages of the rewards, used to smooth the discrepancies, collected from the two training phases. In the figures Fig.~\ref{fig_treino_v} (a) and Fig.~\ref{fig_treino_v} (b) the moving averages of the speeds that led to the rewards obtained during training are presented.

\begin{figure}[htb]
    \begin{minipage}[c][1.1\width]{0.235\textwidth}
	   \centering
	   \includegraphics[width=1.16\textwidth]{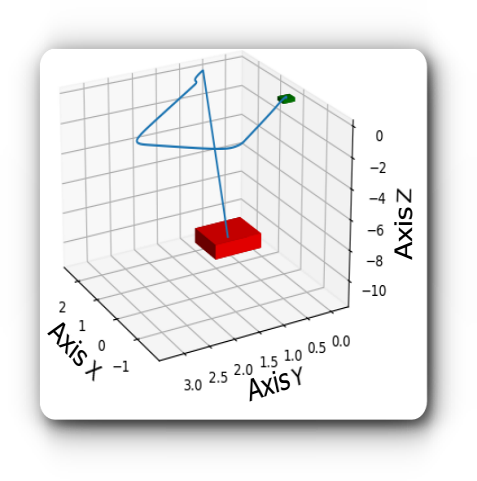}
          \textbf{a)}
	\end{minipage}
    \begin{minipage}[c][1.3\width]{0.235\textwidth}
	   \centering
	   \includegraphics[width=1.16\textwidth]{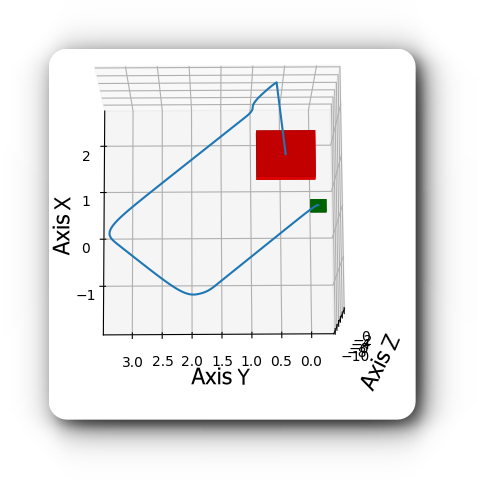}
          \textbf{b)}
	\end{minipage} \vfill
 \caption{a) Side view from learned trajectory b) Top view from learned trajectory.}
      \label{fig_treino_trj}
\end{figure}

During the training period, it was possible to assess that the agent learned to develop trajectories, placing itself in positions similar to the positioning represented by Fig~\ref{fig_treino_trj} (a) and Fig~\ref{fig_treino_trj} (b). This is a consequence of the simultaneous implication of the speeds at each step caused by the search for the greatest accumulation of rewards, aiming at the rewards for reducing the distance and also for the landing. The vehicle learned to adjust to the deformation that occurred in the estimated contour of the target by the effect of the light, which appeared at certain angles taken, and also by the swerves. It was also possible to analyze that the agent generalized the policy in order to position itself placing the target always at the top right of the image with an angle close to $45^0$, leading to critical positions at the point $(0,0)$ of the camera reached in the training periods with random positions.

For better visualization, the collection of the episode, time, and reward obtained during the testing phase is presented. In this way, the maximum rewards obtained in the tests where the landing was performed is shown. Both for critical cases and for others in which the agent's objective was not completed using the $10$ episodes, we show the average value of the reward obtained during the given test. For the time-based graphics, the total time to complete the test is presented.

\subsection{Assessment of the policy learned}
\begin{figure}[htb]
    \begin{minipage}[c][1.1\width]{0.235\textwidth}
	   \centering
	   \includegraphics[width=1.16\textwidth]{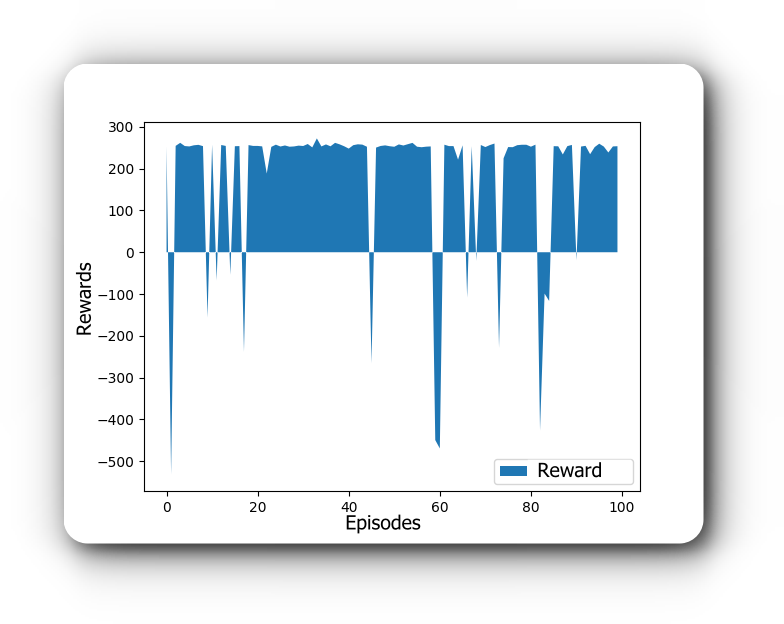}
          \textbf{a)}
	\end{minipage}
    \begin{minipage}[c][1.1\width]{0.235\textwidth}
	   \centering
	   \includegraphics[width=1.16\textwidth]{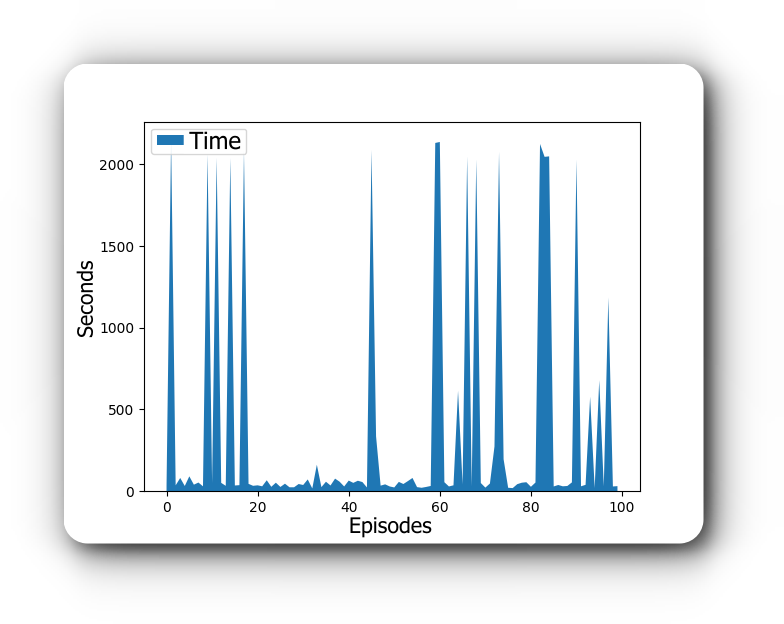}
          \textbf{b)}
	\end{minipage} \vfill
 \caption{Reward and time elapsed during testing phase.}
      \label{fig_teste}
\end{figure}

To validate the policy developed by the agent, parameters similar to those of the training were used, varying only in the number of steps per episode increased up to $2000$ to deal with long trajectories. The positioning of the boat in relation to the UAV was dynamically changed, where random positions and orientations were used within the camera's field of view. Thus, the boat's positions were changed by up to $4m$ from the UAV's position on the $x$ and $y$ axes and could vary both in the positive and negative directions of each axis. To evaluate the quality of the policy, $100$ tests were performed with random positions with the described interval, with a limit of $10$ episodes being defined for the landing to be properly carried out. Thus, the rewards obtained by the agent during each test performed were analyzed, as well as the time required to complete it. In the figures Fig~\ref{fig_teste} (a) and \ref{fig_teste}(b) the rewards and the real-time for carrying out the test are shown, respectively.

\begin{figure}[htb]
    \begin{minipage}[c][1.3\width]{0.235\textwidth}
	   \centering
	   \includegraphics[width=1.16\textwidth]{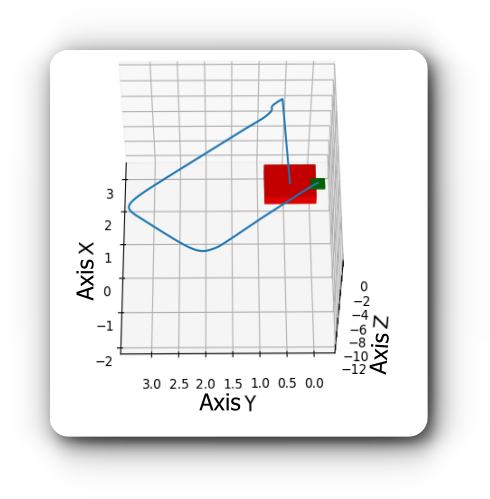}
          \textbf{a)}
	\end{minipage}
    \begin{minipage}[c][1.3\width]{0.235\textwidth}
	   \centering
	   \includegraphics[width=1.16\textwidth]{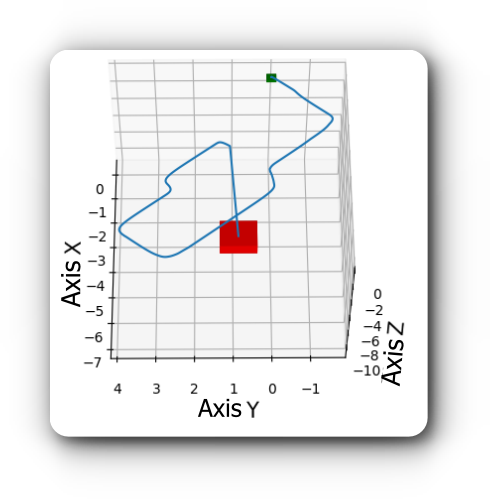}
          \textbf{b)}
	\end{minipage} \vfill
    \begin{minipage}[c][1.3\width]{0.235\textwidth}
	   \centering
	   \includegraphics[width=1.16\textwidth]{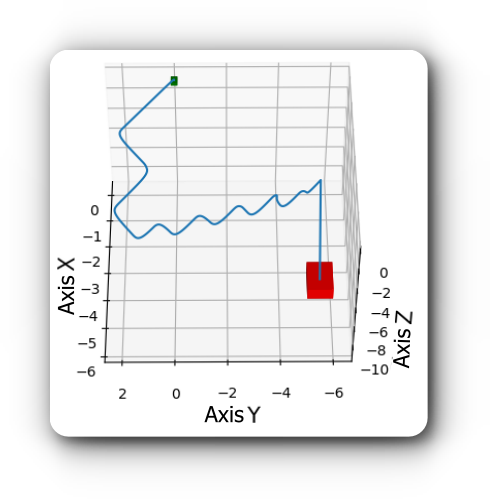}
          \textbf{c)}
	\end{minipage}
    \begin{minipage}[c][1.3\width]{0.235\textwidth}
	   \centering
	   \includegraphics[width=1.16\textwidth]{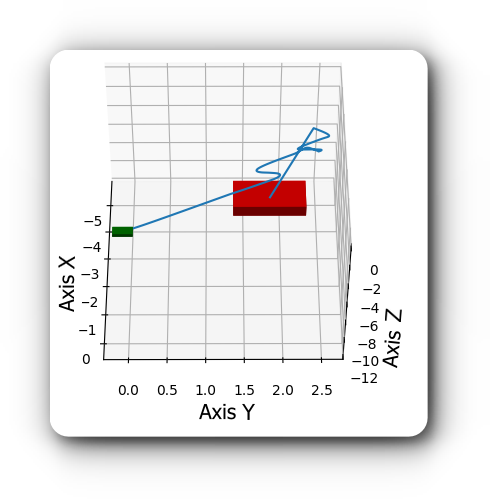}
          \textbf{d)}
	\end{minipage} \vfill
 \caption{Four random positions and orientation sampling running.}
      \label{fig_teste_p}
\end{figure}

The representation of the graphs shows that in most cases the agent managed to make the landing so that the worst rewards characterize the problems generated by the frequent loss of the target. During the training period, this factor could be evaluated and during the tests, it was confirmed, highlighting the problem encountered when landing on targets close to the camera's point $(0.0)$. As a consequence of the policy's generalized trajectory towards the target, the target was lost sight of in these cases. Failure to land in other not-so-critical situations brings a large discrepancy with respect to critical cases. But in terms of runtime, most of the non-convergence cases performed a similar result so they needed the $10$ episodes to complete the test. Analyzing the cases in which it was possible to effectively perform the landing as shown in Fig~\ref{fig_teste_p} (a), Fig~\ref{fig_teste_p} (b) Fig~\ref{fig_teste_p} (c) and Fig~\ref {fig_teste_p} (d), the agent needed an average time equivalent to $30$ seconds to complete the task and then reach the highest reward.

Table~\ref{table_teste_ac} shows the relative and absolute percentages referring to the number of episodes needed to complete the landing and a number of $0$ for tests in which the task was not performed. It can be concluded that the model developed an accuracy equivalent to $85\%$, with most of them achieving a landing in the first episode.
\begin{table}[h]
\caption{Results}
\label{table_teste_ac}
\begin{center}
\begin{tabular}{|c|c|c|}
\hline
Episodes Number & Relative Percentage & Absolute Percentage\\
\hline
$0$ & $0\%$ & $15\%$\\
\hline
$1$ & $100\%$ & $75\%$\\
\hline
$2$ & $50\%$ & $2\%$\\
\hline
$3$ & $33.33\%$ & $2\%$\\
\hline
$4$ & $25\%$ & $1\%$\\
\hline
$6$ & $16.66\%$ & $1\%$\\
\hline

\end{tabular}
\end{center}
\end{table}
\section{Conclusions}
\label{conclusion}

We presented a simple image processing approach through the extraction of features that is functional as a means of feeding \textit{Deep-RL} algorithms to perform tasks actively perceiving the environment. It is capable to deal with noise in the realistic simulation of the aquatic environment, affecting little or nothing in learning with the approach used, resulting in a new light system capable to be embedded in a UAV and perming autonomous landing in boats. 

Future works will be focused in evaluate the system with a real environment operation using a real UAV. Furthermore, other approaches for Deep-RL will be evaluated in the present scenario.

\section*{Acknowledgment}


The authors would like to thank the VersusAI team. This work was partly supported by the CAPES, CNPq and PRH-ANP.

\vspace{-2mm}
\bibliographystyle{./bibliography/IEEEtran}
\bibliography{./bibliography/IEEEabrv,./bibliography/IEEEexample}

\end{document}